\documentclass[10pt,twocolumn,letterpaper]{article}

\usepackage{cvpr}
\usepackage{times}
\usepackage{epsfig}
\usepackage{graphicx}
\usepackage{amsmath}
\usepackage{amssymb}

\usepackage{array}
\usepackage{color}
\usepackage{subfigure}
\usepackage{booktabs}
\usepackage{lipsum}
\usepackage{multirow}
\usepackage[linesnumbered,ruled]{algorithm2e}
\usepackage[font=small,labelsep=period]{caption}

\usepackage[pagebackref=true,breaklinks=true,letterpaper=true,colorlinks,bookmarks=false]{hyperref}

\newcolumntype{P}[1]{>{\centering\arraybackslash}p{#1}}
\newcolumntype{M}[1]{>{\centering\arraybackslash}m{#1}}

\newcommand{\modulename}{style decorator}

\cvprfinalcopy 

\def\cvprPaperID{***} 

\ifcvprfinal\fi
\begin{document}

\title{Avatar-Net: Multi-scale Zero-shot Style Transfer by Feature Decoration}

\author{Lu Sheng$^1$,~~~~Ziyi Lin$^2$,~~~~Jing Shao$^2$,~~~~Xiaogang Wang$^1$\\
$^1$CUHK-SenseTime Joint Lab, The Chinese University of Hong Kong~~~~$^2$SenseTime Research\\
{\tt\small \{lsheng, xgwang\}@ee.cuhk.edu.hk,~~~\{linziyi, shaojing\}@sensetime.com}
}

\twocolumn[{%
\renewcommand\twocolumn[1][]{#1}%
\maketitle
\begin{center}
\centering
\includegraphics[width=\linewidth]{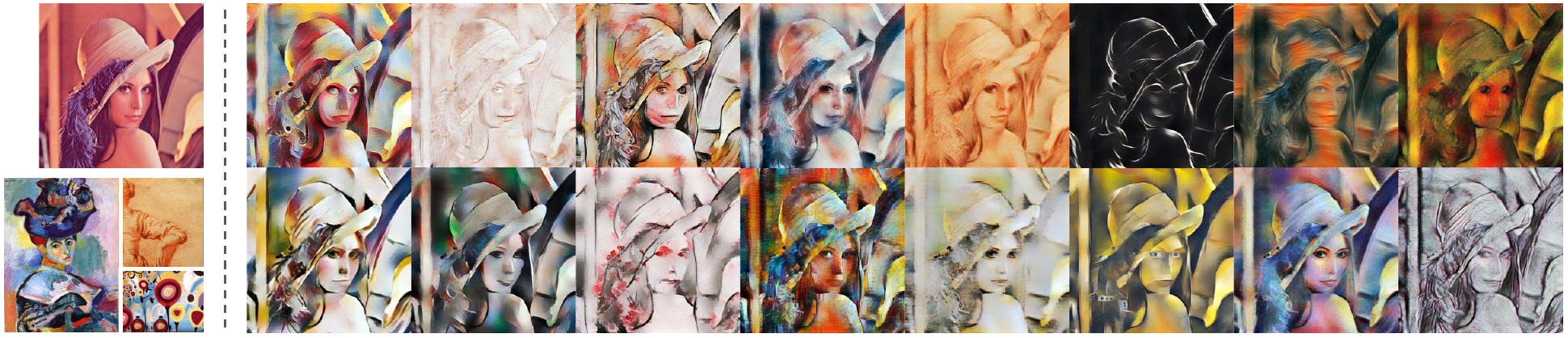}
\captionof{figure}{Exemplar stylized results by the proposed Avatar-Net, which faithfully transfers the \emph{Lena} image by arbitrary style.}
\label{fig:teaser}
\end{center}%
}]



\begin{abstract}

Zero-shot artistic style transfer is an important image synthesis problem aiming at transferring arbitrary style into content images.
However, the trade-off between the generalization and efficiency in existing methods impedes a high quality zero-shot style transfer in real-time.
In this paper, we resolve this dilemma and propose an efficient yet effective Avatar-Net that enables visually plausible multi-scale transfer for arbitrary style.
The key ingredient of our method is a style decorator that makes up the content features by semantically aligned style features from an arbitrary style image, which does not only holistically match their feature distributions but also preserve detailed style patterns in the decorated features.
By embedding this module into an image reconstruction network that fuses multi-scale style abstractions, the Avatar-Net renders multi-scale stylization for any style image in one feed-forward pass.
We demonstrate the state-of-the-art effectiveness and efficiency of the proposed method in generating high-quality stylized images, with a series of applications include multiple style integration, video stylization and etc.

\end{abstract}

\section{Introduction}

Unlike taking days or months to create a particular artistic style by a diligent artist, modern computational methods have enabled fast and reliable style creation for natural images.
Especially inspired by the remarkable representative power of convolutional neural networks (CNNs), the seminal work by Gatys~\etal~\cite{gatys2015texture,gatys2016image} discovered that multi-level feature statistics extracted from a trained CNN notably represent the characteristics of visual styles, which boosts the development of style transfer approaches, either by iterative optimizations~\cite{gatys2016image,li2016combining,wilmot2017stable,li2017Demystifying} or feed-forward networks~\cite{dumoulin2016learned,johnson2016perceptual,li2017diversified,li2017Demystifying,zhang2017multistyle,ulyanov2016texture,wang2017zm,shen2017meta,chen2017stylebank}.
However, a dilemma in terms of generalization and quality versus efficiency hampers the availability of style transfer for arbitrary style.
Since the optimization-based approaches render visually plausible results for various styles at the sacrifice of executing efficiency\cite{gatys2016image,li2016combining,wilmot2017stable}, while the feed-forward networks are either restricted by a finite set of predefined styles~\cite{dumoulin2016learned,johnson2016perceptual,li2017diversified,li2017Demystifying,zhang2017multistyle,ulyanov2016texture,chen2017coherent,chen2018stereoscopic}, or simplify the zero-shot transfer with compromised visual quality~\cite{huang2017arbitrary,shen2017meta,wang2017zm}.

Valuable efforts have been devoted to solving this dilemma.
A common practice is employing external style signals to supervise the content modification~\cite{huang2017arbitrary,shen2017meta,zhang2017multistyle,wang2017zm,chen2017stylebank} on a feed-forward network.
But these methods require the networks trained by the perceptual loss~\cite{johnson2016perceptual}, which has been known instable and produces compromised stylized patterns~\cite{gupta2017characterizing,wilmot2017stable}.
In contrast, another category~\cite{li2017universal,chen2016fast} manipulates the content features under the guidance of the style features in a shared high-level feature space.
By decoding the manipulated features back into the image space with a style-agnostic image decoder, the reconstructed images will be stylized with seamless integration of the style patterns.
However, current techniques may either over-distort the content and add unconstrained patterns~\cite{li2017universal}, or fail to retrieve complete style patterns when large domain gap exists between the content and style features~\cite{chen2016fast}.

Facing the aforementioned challenges, we propose a zero-shot style transfer model \emph{Avatar-Net} that follows a similar style-agnostic paradigm~\cite{li2017universal,chen2016fast} in the aforementioned latter category.
Specifically, we introduce a novel patch-based \emph{style decorator} module that decorates the content features with the characteristics of the style patterns, while keeps the content semantically perceptible.
The style decorator does not only match the holistic style distribution, but also explicitly retrieves detailed style patterns without distortion.
In the meanwhile, other than auto-encoders~\cite{chen2016fast,li2017universal,chen2017stylebank} for semantic feature extraction and style-agnostic image decoder, we introduce an innovative hourglass network with multiple skip connections for multi-scale holistic style adaptation.
It is thus straightforward to perform multi-scale zero-shot style transfer by 1) extracting the content and style features via its encoding module, 2) decorating the content features by the patch-based style decorator, and 3) progressively decoding the stylized features with multi-scale holistic style adaptation.
Note that the proposed hourglass network is trained solely to reconstruct natural images.
As shown in Fig.~\ref{fig:teaser}, the proposed Avatar-Net can synthesize visually plausible stylized \emph{Lena} images for arbitrary style.
Comprehensive evaluations have been conducted to compare with the prior style transfer methods~\cite{huang2017arbitrary,chen2016fast,li2017universal,johnson2016perceptual}, our method achieves the state-of-the-art performance in terms of both visual quality and efficiency.

Our contributions in this paper are threefold:
(1) A novel patch-based feature manipulation module named as style decorator, transfers the content features to semantically nearest style features and simultaneously minimizes the discrepancy between their holistic feature distributions.
(2) A new hourglass network equipped with multi-scale style adaptation enables visually plausible multi-scale transfer for arbitrary style in one feed-forward pass. 
(3) Theoretical analysis proves that the style decorator module owns superior transferring ability, and the experimental results demonstrate the effectiveness of the proposed method with superior visual quality and economical processing cost.

\section{Related Work} 
\label{sec:related_work}

Style transfer is a kind of non-realistic rendering techniques~\cite{kyprianidis2013state} that is closely related to texture synthesis~\cite{efros2001image,elad2017style}.
It usually exploits local statistics for efficient cross-view dense correspondences and texture quilting~\cite{hertzmann2001image,efros2001image,shih2014style,liao2017visual}.
Although these methods produce appealing stylized images~\cite{shih2014style,liao2017visual}, dense correspondences are limited to a pair of images with similar contents, thus inapplicable to zero-shot style transfer.

\vspace{+0.5mm}
\noindent\textbf{Optimization-based Stylization.}
Gatys~\etal~\cite{gatys2016image,gatys2015texture} at the first time formulated the style as multi-level feature correlations (\ie, Gram matrix) from a trained neural network for image classification, and defined the style transfer as an iterative optimization problem that balances the content similarity and style affinity in the feature level (or termed as perceptual loss~\cite{johnson2016perceptual}).
A number of variants have been developed thereafter to adapt this framework to different scenarios and requirements, including photorealistic rendering~\cite{luan2017deep}, semantically composite transfer~\cite{champandard2016semantic}, temporal coherence~\cite{ruder2016artistic} and so on.
Changing the global style statistics into Markovian feature assembling, a similar framework is also proposed for semantic image synthesis~\cite{li2016combining}.
Despite visual appealing performances for arbitrary style, the results are not stable~\cite{ruder2016artistic,wilmot2017stable} and the perceptual loss usually requires careful parameter tunning for different styles~\cite{gupta2017characterizing}.
Moreover, the inefficiency of the optimization-based approaches restrains real-time applications.

\noindent\textbf{Feed-Forward Approximation.}
Recent researches~\cite{johnson2016perceptual,ulyanov2016texture,li2016precomputed,zhang2017multistyle,chen2017stylebank,shen2017meta,wang2017zm,li2017diversified,dumoulin2016learned} try to tackle the complexity issue by approximating the iterative back-propagating procedure to feed-forward neural networks, either trained by the perceptual loss or Markovian generative adversarial loss~\cite{li2016precomputed}.
However, \cite{johnson2016perceptual,ulyanov2016texture,li2016precomputed} have to train a independent network for every style.
To strengthen the representation power for multiple styles, StyleBank~\cite{chen2017stylebank} learns filtering kernels for styles and Li~\etal~\cite{li2017diversified} transfered styles by binary selection units, as well as Dumoulin~\etal proposed conditional instance normalization~\cite{dumoulin2016learned} controlled by channel-wise statistics learned for each style.
But the manually designed style abstractions are often short for the representation of unseen styles and the combinational optimization over multiple styles often compromises the rendering quality~\cite{li2017Demystifying,gupta2017characterizing}.

\noindent\textbf{Zero-shot Style Transfer.}
To achieve zero-shot style transfer in a feed-forward network, Huang~\etal~\cite{huang2017arbitrary} adjusted channel-wise statistics of the content features by adaptive instance normalization (AdaIN), and trained a feature decoder by a combinational scale-adapted content and style losses.
Chen~\etal~\cite{chen2016fast} swapped the content feature patches with the closest style features (\ie Style-Swap) at the intermediate layer of a trained auto-encoder, while Li~\etal~\cite{li2017universal} transferred multi-level style patterns by recursively applying whitening and coloring transformation (WCT) to a set of trained auto-encoders with different levels.
However, AdaIN over-simplifies the transferring procedure and Style-Swap cannot parse sufficient style features when content and style images are semantically different.
WCT is the state-of-the-art method, but the holistic transformation may results in distorted style patterns in the transferred features, and generate unwanted patterns in the output image.

The proposed zero-shot style transfer is related to~\cite{li2017universal,chen2016fast,huang2017arbitrary}.
But the \modulename{} improves AdaIN and WCT as it reserves the detailed style patterns rather than the parameterized feature statistics.
It also outperforms Style-Swap as it effectively parses the complete set of style features regardless of their domain gap.
Moreover, the proposed network performs multi-scale style adaptation in one feed-forward pass, which surpasses AdaIN and WCT since AdaIN requires a style-oriented image decoder and WCT needs a set of recursive feed-forward passes to enable multi-scale style transfer.

\section{Style Transfer via Feature Manipulation}
\label{sec:non_parametric_feature_transfer}

Let a trained feed-forward network extract the features in the bottleneck layer for an image $\mathbf{x}$ as $\mathbf{z} = E_{\boldsymbol\theta_\text{enc}}(\mathbf{x}) \in \mathbb{R}^{H\times W\times C}$, with height $H$, width $W$ and channel $C$.
The extracted features are decoded via $\tilde{\mathbf{x}} = D_{\boldsymbol\theta_\text{dec}}(\mathbf{z})$ to the reconstructed images.
The encoder and decoder modules are parameterized by $\boldsymbol\theta_\text{enc}$ and $\boldsymbol\theta_\text{dec}$, respectively.

Denote $\mathbf{z}_c$ and $\mathbf{z}_s$ as the features for the content image $\mathbf{x}_c$ and style image $\mathbf{x}_s$.
By transferring $\mathbf{z}_c$ into the domain of $\mathbf{z}_s$, it is desired that the transferred features $\mathbf{z}_{cs} = \mathcal{F}(\mathbf{z}_{c}; \mathbf{z}_s)$ infer the spatial distribution of the content features and the textural characteristics of the style features.
The key challenge is to design a feature transfer module that both holistically adapts the domain of $\mathbf{z}_{cs}$ to that of $\mathbf{z}_s$ and semantically links elements in $\mathbf{z}_{cs}$ to paired elements in $\mathbf{z}_s$ and $\mathbf{z}_c$.

\subsection{Revisiting Normalized Cross-correlation}
\label{sub:revisiting_normalized_cross_correlation}

A type of modules generates $\mathbf{z}_{cs}$ by non-parametrically swapping the patches in $\mathbf{z}_c$ by their nearest neighbors in $\mathbf{z}_s$.
It encourages concrete style patterns in $\mathbf{z}_{cs}$ and explicitly binds $\mathbf{z}_c$ and $\mathbf{z}_{cs}$ by their spatial distributions.

\vspace{+1mm}
\noindent\textbf{Normalized Cross-correlation (NCC).}
It is one effective metric for patch-wise nearest neighbor searching~\cite{li2016combining,chen2016fast}, by scoring the cosine distance between a content patch $\phi_i(\mathbf{z}_c)$ and a style patch $\phi_j(\mathbf{z}_s)$ and returning the nearest patch
\begin{equation*}
\phi_{i}(\mathbf{z}_{cs}) = \underset{j\in\{1,\ldots,N_s\}}{\arg\max} \frac{\langle\phi_i(\mathbf{z}_c), \phi_j(\mathbf{z}_s)\rangle}{\|\phi_i(\mathbf{z}_c)\| \|\phi_j(\mathbf{z}_s)\| }, i\in\{1,\ldots,N_c\}.
\end{equation*}
However, each $\phi_i(\mathbf{z}_c)$ is required to compare with any patch in $\mathbf{z}_s$.
Thanks to the GPU accelerated convolution operations, NCC can be converted into several steps of efficient convolutions, such as the way applied in Style-Swap~\cite{chen2016fast,li2016combining}.

\begin{figure}
\centering
\includegraphics[width=\linewidth]{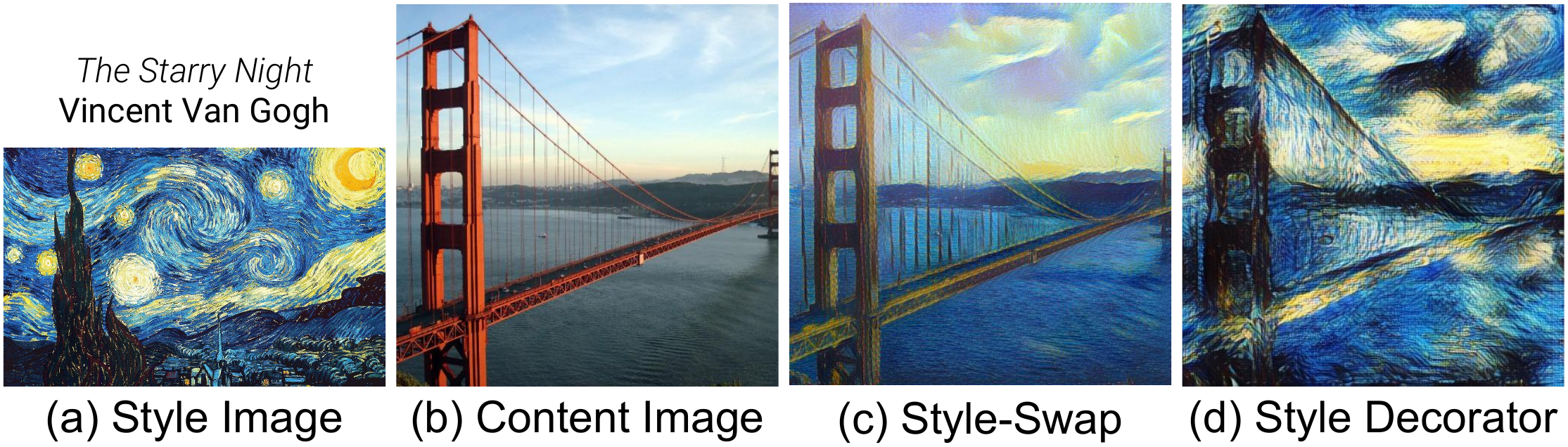}
\caption{Style patterns are overlaid onto the content patches according to their semantic similarity rather than their texture agreement. (a) and (b) are the style and content images. (c)-(d) are the results by Style-Swap~\cite{chen2016fast} and the proposed style decorator. Style-Swap pays too much attention on the textural similarity, thus receives fewer high-level style patterns.}
\label{fig:patch_match_comparison}
\end{figure}

\begin{figure*}
\centering
\includegraphics[width=\linewidth]{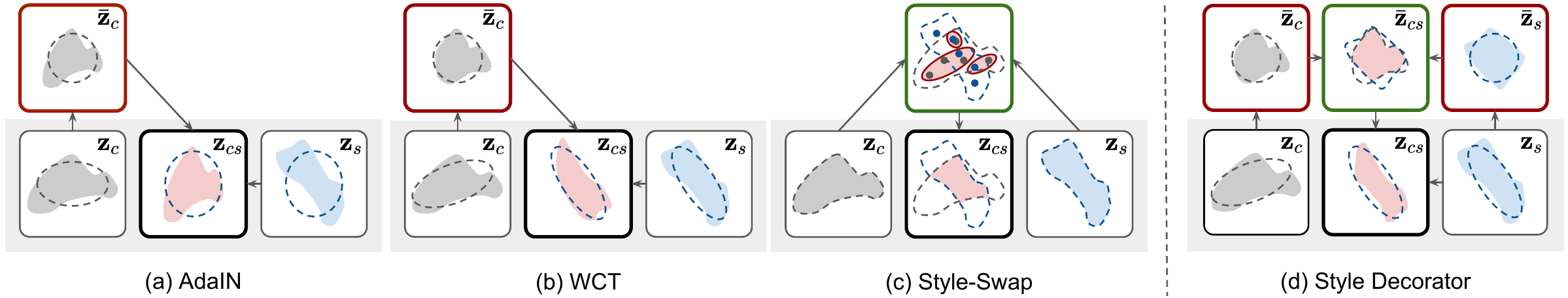}
\caption{Comparison of distribution transformation by different feature transfer modules. (a) AdaIN cannot completely dispel the textures from $\mathbf{z}_c$ and propagate enough style textures to $\mathbf{z}_{cs}$. (b) WCT has optimal matching between $\mathcal{G}(\mathbf{z}_{cs})$ and $\mathcal{G}(\mathbf{z}_s)$ but it introduces rare patterns observed in the transformed features. (c) Style-Swap matches patches from $\mathbf{z}_c$ and $\mathbf{z}_s$, and thus only captures the overlapped features. (d) Style decorator matches the normalized features $\bar{\mathbf{z}}_{c}$ and $\bar{\mathbf{z}}_{s}$, the generated $\mathbf{z}_{cs}$ parses all possible style patterns to the content features. The red boxes show the normalization and the green boxes visualize the normalized cross-correlation.}
\label{fig:feature_coverage_comparison}
\end{figure*}

\vspace{+1mm}
\noindent\textbf{Discussions.}
Although concrete style patterns are preserved in $\mathbf{z}_{cs}$, it is highly biased towards the content features in $\mathbf{z}_c$, since it suggests the matched style patches strictly follow the local variations as the content patches.
Therefore, the spiral patterns in sky in Vincent Van Gogh's \emph{starry night} cannot be propagated to a sky patch in the content image, as shown in Fig.~\ref{fig:patch_match_comparison}.
And only a limited portion of style patterns are parsed to assemble $\mathbf{z}_{cs}$ when the domain of $\mathbf{z}_c$ and that of $\mathbf{z}_s$ are far apart, as depicted by the swapping procedure shown in Fig.~\ref{fig:feature_coverage_comparison}(c).
Thus the stylized images preserve more content patterns with only a small portion of semantically aligned style patterns.
In fact, two patches can be matched as long as they are paired in some forms of semantic agreement rather than this over-restrictive patch similarity.
It means that a desired patch matching can be \emph{invariant} to their domain gap and the patches are paired by comparing their local statistics in a \emph{common} space that the textural characteristics have been dispelled.

\subsection{Style Decorator}
\label{sub:feature_transfer_module}

To resolve drawbacks raised from the existing feature transfer modules, we propose a novel \modulename{} module that robustly propagates the thorough style patterns onto the content features so that the distributions of $\mathbf{z}_{cs}$ and $\mathbf{z}_s$ are maximally aligned and the detailed style patterns are semantically perceptible in $\mathbf{z}_{cs}$.
It is achieved by a relaxed cross correlation between patches in a normalized coordinate that sufficiently whitens the textures from the original feature spaces for both features.
The swapped normalized features $\bar{\mathbf{z}}_{cs}$ explicitly correspond to the original style features, thus from which we are able to reconstruct the stylized features $\mathbf{z}_{cs}$ that both statistically follow the style of $\mathbf{z}_s$ and the spatial layout of the content feature $\mathbf{z}_c$.
The style decorator is achieved by a three-step procedure:

\vspace{+1mm}
\noindent\textbf{Step-I: Projection.}
We project $\mathbf{z}_c$ and $\mathbf{z}_s$ onto the same space, written as $\bar{\mathbf{z}}_c$ and $\bar{\mathbf{z}}_s$, with sufficient compactness and similar scale in magnitude.
In detail, the projection operation for the content (style) feature is
\begin{equation}
\bar{\mathbf{z}}_c = \mathbf{W}_c\otimes(\mathbf{z}_c - \boldsymbol\mu(\mathbf{z}_c)), \bar{\mathbf{z}}_s = \mathbf{W}_s\otimes(\mathbf{z}_s - \boldsymbol\mu(\mathbf{z}_s)),
\end{equation}
where $\otimes$ denotes convolutional operator, and the whitening kernels $\mathbf{W}_c$ and $\mathbf{W}_s$ are derived by some forms of whitening matrices onto the covariance matrices $\mathcal{C}(\mathbf{z}_c)$ and $\mathcal{C}(\mathbf{z}_s)$.
And $\boldsymbol\mu(\mathbf{z}_c)$ and $\boldsymbol\mu(\mathbf{z}_s)$ are the mean features.
$\bar{\mathbf{z}}_c$ and $\bar{\mathbf{z}}_s$ keep the characteristics from their original data, but both of them statistically follow standard normal distribution.

\begin{figure}
\centering
\includegraphics[width=\linewidth]{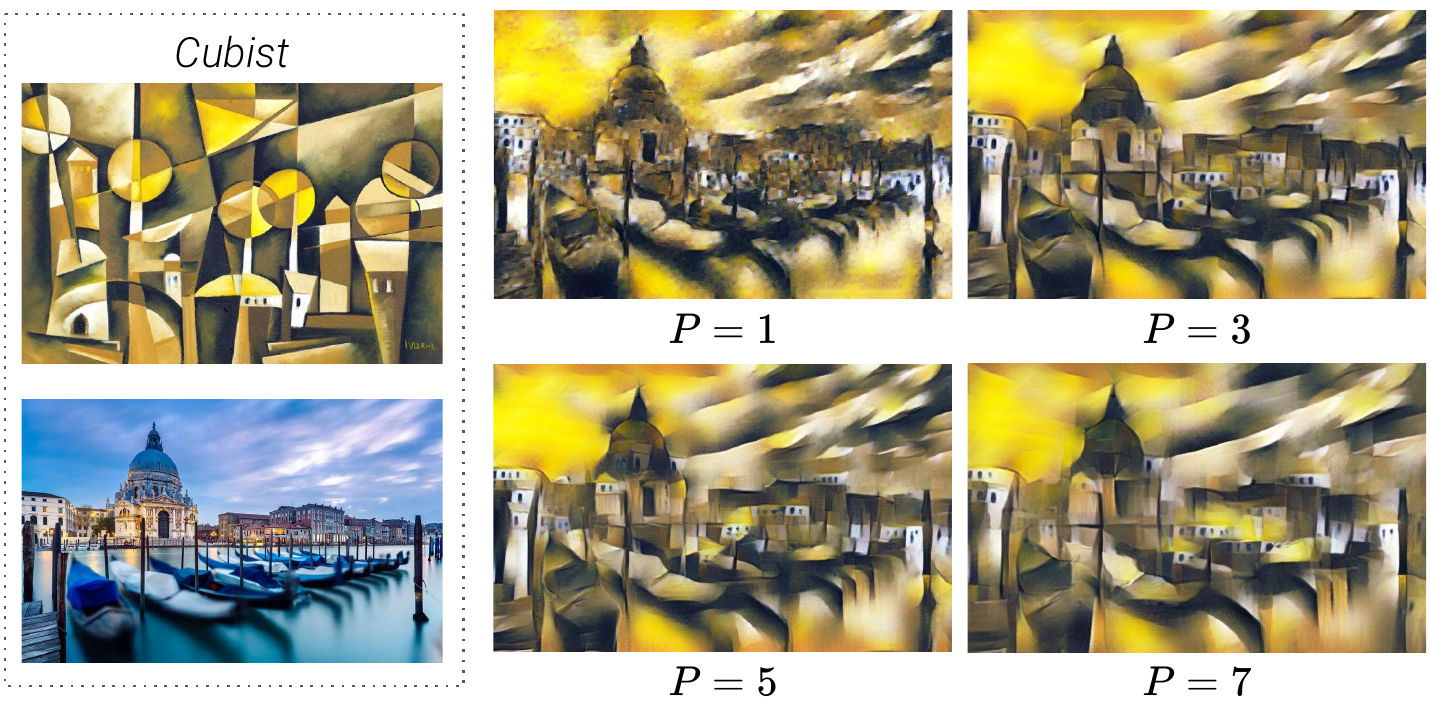}
\caption{Stylization controlled by different patch size. With an increasing patch size, the stylized image tends to be more blocky like the geometric patterns in the style image \emph{cubist}.}
\label{fig:patch_size_comparison}
\end{figure}

\begin{figure}
\centering
\includegraphics[width=\linewidth]{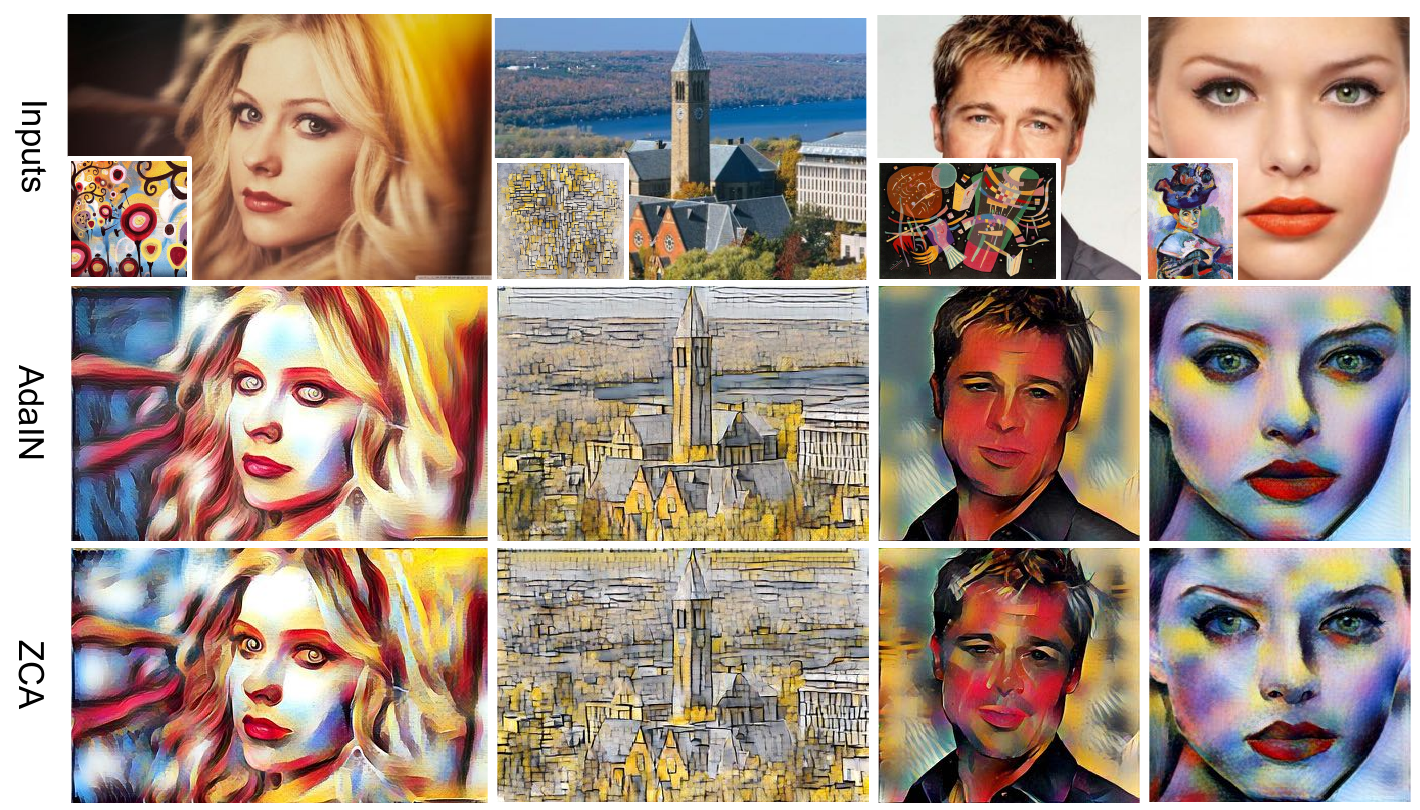}
\caption{Both ZCA and AdaIN are effective for the projection and reconstruction steps. AdaIN will preserve a bit more amount of content patterns but it is much more efficient than ZCA.}
\label{fig:different_wct_modules}
\end{figure}

\begin{figure*}
\centering
\includegraphics[width=\linewidth]{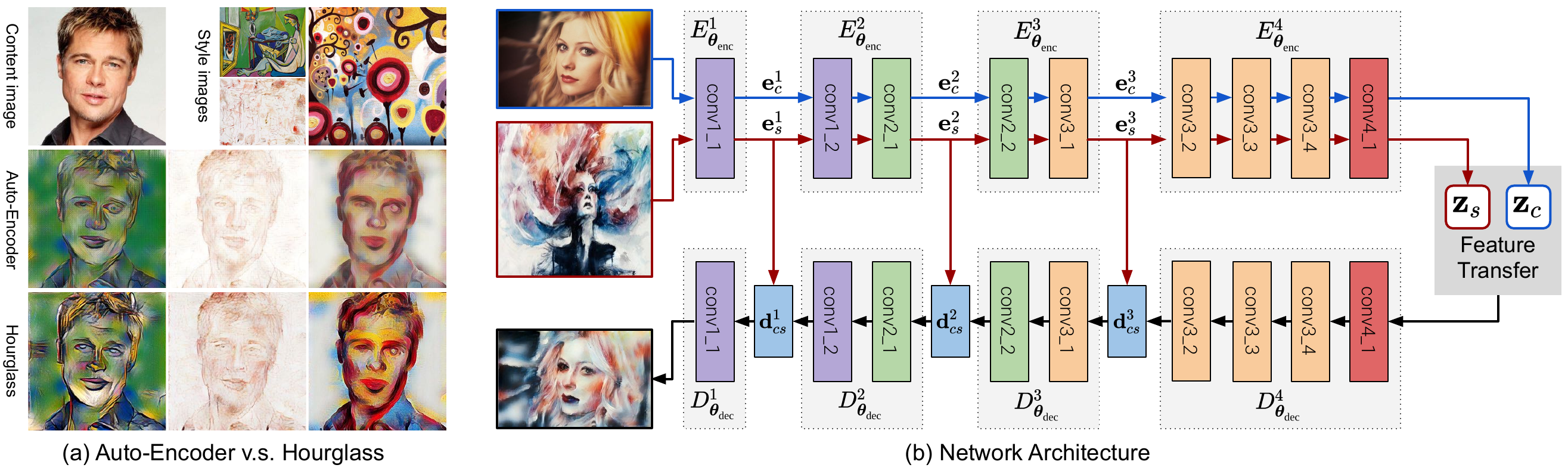}
\caption{(a) Stylization comparison by auto-encoder and style-augmented hourglass networks. Style decorator is applied as the feature transfer module. The auto-encoder and the hourglass networks share the same main branch. (b) The network architecture of the proposed style-augmented hourglass network. Detailed implementation is depicted in Sec.~\ref{sub:network_architecture_and_training}.}
\label{fig:network_architecture}
\end{figure*}

\vspace{+1mm}
\noindent\textbf{Step-II: Matching and Reassembling.}
In this normalized domain, we want to align any element in $\bar{\mathbf{z}}_c$ with the nearest element in $\bar{\mathbf{z}}_s$, so as to reconstruct $\bar{\mathbf{z}}_c$ by reassembling the corresponded elements in $\bar{\mathbf{z}}_s$.
In this case, the reassembled normalized style features $\bar{\mathbf{z}}_{cs}$ contains concrete normalized style patterns but their spatial layouts come from the normalized content features.
Therefore, the matching and reassembling operations seamlessly incorporate the desired content and style characteristics in the resultant $\bar{\mathbf{z}}_{cs}$ .

We apply the patch matching and reassembling via the help of normalized cross-correlation, which can be efficiently transformed into several convolutional operations as
\begin{equation}
\bar{\mathbf{z}}_{cs} = \boldsymbol\Phi(\bar{\mathbf{z}}_s)^\top \otimes \mathcal{B}(\bar{\boldsymbol\Phi}(\bar{\mathbf{z}}_s)\otimes\bar{\mathbf{z}}_c),
\end{equation}
where $\boldsymbol\Phi(\bar{\mathbf{z}}_s)\in\mathbb{R}^{P\times P \times C \times (H\times W)}$ is the style kernel consists of patches in $\bar{\mathbf{z}}_s$, and $P$ is the patch size.
In addition, $\bar{\boldsymbol\Phi}(\bar{\mathbf{z}}_s)$ is the normalized style kernel that is divided by the channel-wise $\ell_2$ norm of each patch.
The patch matching procedure is at first convolving the normalized content feature $\bar{\mathbf{z}}_c$ with $\bar{\boldsymbol\Phi}(\bar{\mathbf{z}}_s)$ and then binarizing the resultant scores (\ie, $\mathcal{B}(\cdot)$) such that the maximum value along the channel is $1$ and the rest are $0$. 
Then the reassembling procedure applies a deconvolutional operation by $\boldsymbol\Phi(\bar{\mathbf{z}}_s)^\top$ to regenerate the style patterns from the binary scores.

This patch matching between $\bar{\mathbf{z}}_c$ and $\bar{\mathbf{z}}_s$ parses effective and complete correspondences since $\bar{\mathbf{z}}_c$ and $\bar{\mathbf{z}}_s$ have the maximal overlap with each other, thus each element in $\bar{\mathbf{z}}_c$ can find a suitable correspondence in $\bar{\mathbf{z}}_s$ whilst possibly every element in $\bar{\mathbf{z}}_s$ can be well retrieved, as demonstrated in Fig.~\ref{fig:feature_coverage_comparison}(d).
The matching and reassembling step actually adds two style-controlled convolutional layers to the feed-forward network and thus its implementation is usually efficient in modern deep learning platforms.

\vspace{+1mm}
\noindent\textbf{Step-III: Reconstruction.}
After reorganizing the normalized style patches into $\bar{\mathbf{z}}_{cs}$, we will reconstruct it into the domain of the style feature $\mathbf{z}_s$.
In detail, we apply the coloring transformation with respect to $\mathbf{z}_s$ into the reassembled feature, the reconstruction of the stylized feature is
\begin{equation}
\mathbf{z}_{cs} = \mathbf{C}_s \otimes \bar{\mathbf{z}}_{cs} + \boldsymbol\mu(\mathbf{z}_s),
\end{equation}
where the coloring kernel $\mathbf{C}_s$ is also derived from the covariance matrix $\mathcal{C}(\mathbf{z}_s)$.

\vspace{+1mm}
\noindent\textbf{Discussions.}
Similar as WCT and AdaIN, the \textit{projection} and \textit{reconstruction} steps are designed to encourage that the second-order statistics of the stylized feature $\mathbf{z}_{cs}$ to be matched to those of the style feature $\mathbf{z}_s$\footnote{They will not be exactly the same, since the population of $\bar{\mathbf{z}}_{cs}$ has been changed to follow a similar spatial distribution of $\bar{\mathbf{z}}_c$.}, \ie, their Gram matrices $\mathcal{G}(\mathbf{z}_s) \simeq \mathcal{G}(\mathbf{z}_{cs})$.
Actually, we may employ Adaptive Instance Normalization (AdaIN)~\cite{huang2017arbitrary}, or Zero-phase Component Analysis (ZCA)~\cite{kessy2017optimal} used in WCT~\cite{li2017universal} as the whitening and coloring kernels.
Whitened features $\bar{\mathbf{z}}$ by ZCA minimize the $\ell_2$ distance with respect to $\mathbf{z}$, thus they marginally preserve the original feature orientations and element-wise distributions.
AdaIN is much faster than ZCA but it cannot optimally dispel feature correlations as what ZCA does, thus the transferred features contain a slightly more content patterns, as shown in Fig.~\ref{fig:feature_coverage_comparison}(a) and (b).
But for the sake of efficiency, the AdaIN induced style decorator is good enough for a real-time zero-shot style transfer, as shown in Fig.~\ref{fig:different_wct_modules}.

On the other hand, the \textit{matching and reassembling} step pairs the elements in $\mathbf{z}_c$ and $\mathbf{z}_s$ by matching their normalized counterparts, thus it effectively reduces the bias and enriches the parsing diversity, which is much more effective than the way by Style-Swap~\cite{chen2016fast}.
In addition, the patch size $P$ affects the scale of the presented styles since a larger patch size leads to a more global style patterns, for example the blocky geometric patterns of style \emph{cubist} in Fig.~\ref{fig:patch_size_comparison}.

\section{Multi-scale Zero-shot Style Transfer}
\label{sec:multi_scale_zero_shot_style_transfer}

The proposed Avatar-Net employs a hourglass network with multi-scale style adaptation modules that progressively fuse the styles from the encoded features into the corresponded decoded features, thus it enables multi-scale style transfer in one feed-forward pass, as shown in Fig.~\ref{fig:network_architecture}(a).

As shown in Fig.~\ref{fig:network_architecture}(b), the main branch of the Avatar-Net is analogous to conventional encoder-decoder architectures that stack an encoder $E_{\boldsymbol\theta_\text{enc}}(\cdot)$ and an decoder $D_{\boldsymbol\theta_\text{dec}}(\cdot)$.
In detail, the encoder is a concatenation of several encoding blocks $E_{\boldsymbol\theta_\text{enc}}^l(\cdot)$ that progressively extracts the intermediate features $\mathbf{e}^l = E_{\boldsymbol\theta_\text{enc}}^l(\mathbf{e}^{l-1}), l \in \{1, \ldots, L\}$ and produces the bottleneck feature $\mathbf{z}$ after the final block $\mathbf{z} = E_{\boldsymbol\theta_\text{enc}}^{L+1}(\mathbf{e}^{L})$.
Inversely, the decoder progressively generates intermediate features $\mathbf{d}^l = D^{l+1}_{\boldsymbol\theta_\text{dec}}(\mathbf{d}^{l+1})$ starting from $\mathbf{z}$, in which $\mathbf{d}^l$ is further updated by fusing with the corresponded encoded features $\mathbf{e}^l$ via the style adaptive feature fusion module.
The output image is decoded by $\tilde{\mathbf{x}} = D_{\boldsymbol\theta_\text{dec}}^1(\mathbf{d}^1)$.

\begin{figure*}
\centering
\includegraphics[width=\linewidth]{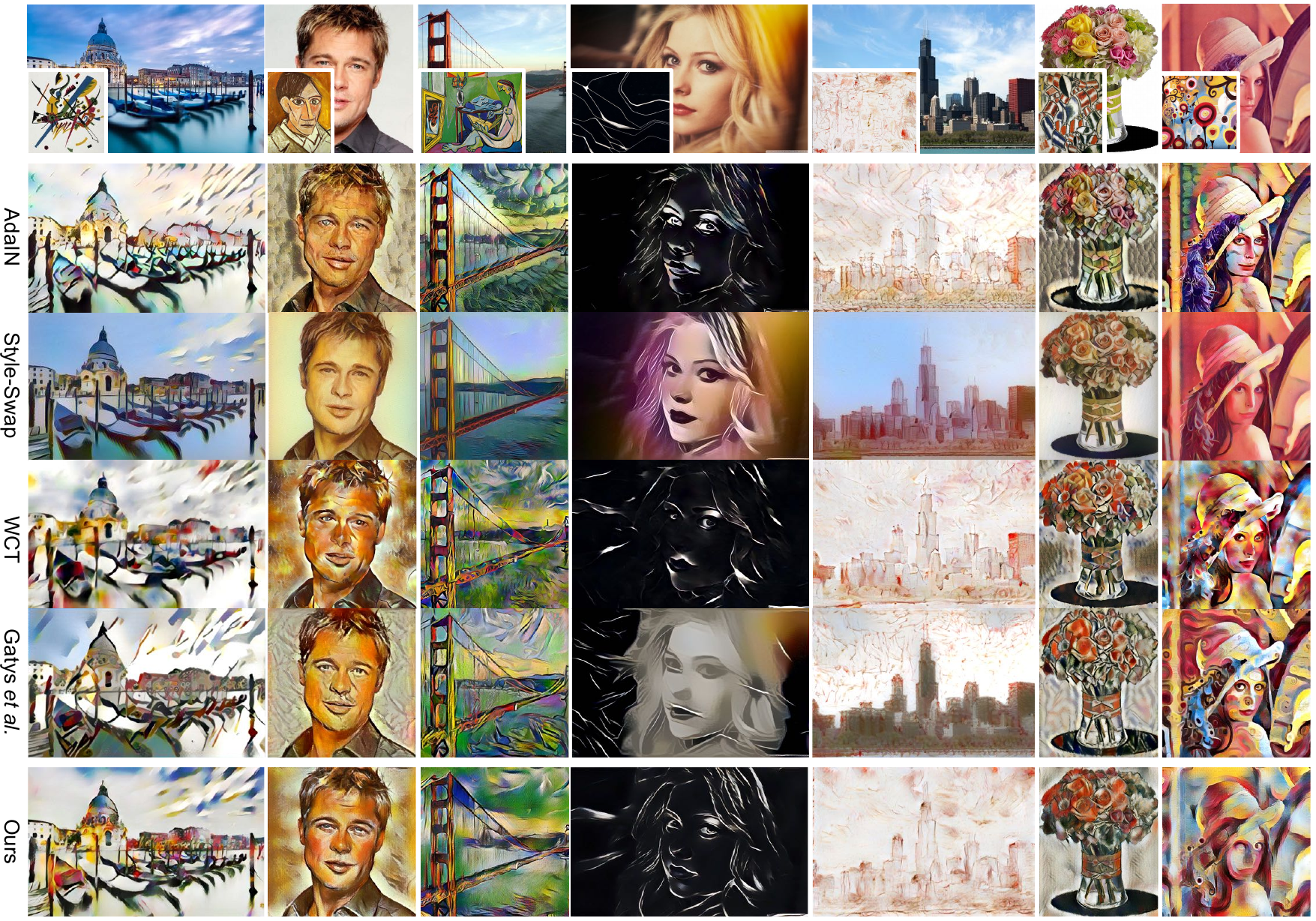}
\caption{Comparison of some exemplar stylization results. The top row shows the content and style pairs and the rest rows present the stylized results by AdaIN~\cite{huang2017arbitrary}, Style-Swap~\cite{chen2016fast}, WCT~\cite{li2017universal}, Gatys~\etal~\cite{gatys2016image}, and the proposed method.}
\label{fig:style_transfer_comparison}
\end{figure*}

Given the set of encoded style features and the decoded stylized feature pairs $(\mathbf{e}_s^l, \mathbf{d}_{cs}^l), l\in \{ 1,\ldots,L\}$ extracted from the main branch of the Hourglass network, the Style Fusion module is similarly as AdaIN~\cite{huang2017arbitrary}:
\begin{equation}
\mathcal{F}_\text{SF}(\mathbf{d}_{cs}^l; \mathbf{e}_s^l) = \boldsymbol\sigma(\mathbf{e}_s^l) \circ \left( \frac{\mathbf{d}_{cs}^l-\boldsymbol\mu(\mathbf{d}_{cs}^l)}{\boldsymbol\sigma(\mathbf{d}_{cs}^l)} \right) + \boldsymbol\mu(\mathbf{e}_s^l),
\end{equation}
where $\circ$ denotes channel-wise multiplication and $\boldsymbol\sigma(\cdot)$ is the channel-wise standard deviation.
One may argue that this module is suboptimal to ZCA as it does not optimally match their second-order statistics, as visualized in Fig.~\ref{fig:feature_coverage_comparison}(a).
But as the proposed network suggest similar pattern distribution of $\mathbf{d}_{cs}^l$ as that of the encoded features $\mathbf{e}_s^l$, the training of the decoder will let the proposed module move towards ZCA.
Moreover, its economical computational complexity also speeds up the stylization.

The style transfer requires a special procedure to fit the proposed network architecture.
At first, Avatar-Net takes a content image $\mathbf{x}_c$ and an arbitrary style image $\mathbf{x}_s$ as inputs, and extracts $\mathbf{z}_c$ and $\mathbf{z}_s$ in the bottleneck layer through the encoder module.
In the meanwhile, the style image $\mathbf{x}_s$ also bypasses the multi-scale encoded features $\{\mathbf{e}^l_s\}_{l=1}^L$.
Secondly, the content feature $\mathbf{z}_c$ is then transferred based on the style features $\mathbf{z}_s$ through the proposed style decorator module.
In the end, the stylized image $\tilde{\mathbf{x}}_{cs}$ is inverted by the decoded module $D_{\boldsymbol\theta_\text{dec}}(\mathbf{z}_{cs})$ with multiple style fusion modules that progressively modify the the decoded features $\mathbf{d}_{cs}^l$ under the guidance of multi-scale style patterns $\mathbf{e}_s^l$, from $l = L$ to $1$, as shown in Fig.~\ref{fig:network_architecture}(b).

\section{Experimental Results and Discussions}
\label{sec:experimental_results_and_discussions}

\subsection{Network Architecture And Training}
\label{sub:network_architecture_and_training}

Avatar-Net copies the architecture of a pretrained VGG-19 (up to \texttt{conv4\_1})~\cite{simonyan2014very} to the encoder $E_{\boldsymbol\theta_\text{enc}}(\cdot)$.
The decoder $D_{\boldsymbol\theta_\text{dec}}(\cdot)$ is randomly initialized and mirrors the encoder with all pooling layers replaced by nearest upsampling and padding layers by reflectance padding.
The shortcut connections link \texttt{conv1\_1}, \texttt{conv2\_1} and \texttt{conv3\_1} to their corresponded decoded layers.
Normalization is not applied in each \texttt{conv} layer so as to increase the reconstruction performance and stability, as suggested by~\cite{huang2017arbitrary}.

The proposed model is to reconstruct perceptually similar images as the input images~\cite{johnson2016perceptual}:
\begin{multline}
\ell_\text{total} = \| \mathbf{x} - \tilde{\mathbf{x}} \|_2^2 + \\ \lambda_{1} \frac{1}{|\mathcal{I}|}\sum_{i\in\mathcal{I}} \| \boldsymbol\Psi_\text{VGG}^i(\mathbf{x}) - \boldsymbol\Psi_\text{VGG}^i(\tilde{\mathbf{x}}) \|_2^2 + \lambda_2 \ell_\text{TV}(\mathbf{x}),
\end{multline}
where $\boldsymbol\Psi_\text{VGG}^i(\mathbf{x})$ extract the feature of $\mathbf{x}$ at layer $i$ in a fixed VGG-19 network.
The set $\mathcal{I}$ contains \texttt{conv1\_1}, \texttt{conv2\_1}, \texttt{conv3\_1} and \texttt{conv4\_1} layers.
Total variation loss $\ell_\text{TV}(\mathbf{x})$ is also added to enforce piece-wise smoothness.
The weighting parameters are simply set as $\lambda_1 = 0.1$ and $\lambda_2 = 1$ for balancing the gradients from each term.

We train the network on the MSCOCO~\cite{lin2014microsoft} dataset with roughly $80,000$ training samples.
Adam optimizer is applied with a fixed learning rate of $0.001$ and a batch size of $16$.
During the training phase, the training images are randomly resized and cropped to $256\times256$ patches.
Note that our method is suitable for images with arbitrary size.

\begin{figure*}
\centering
\includegraphics[width=\linewidth]{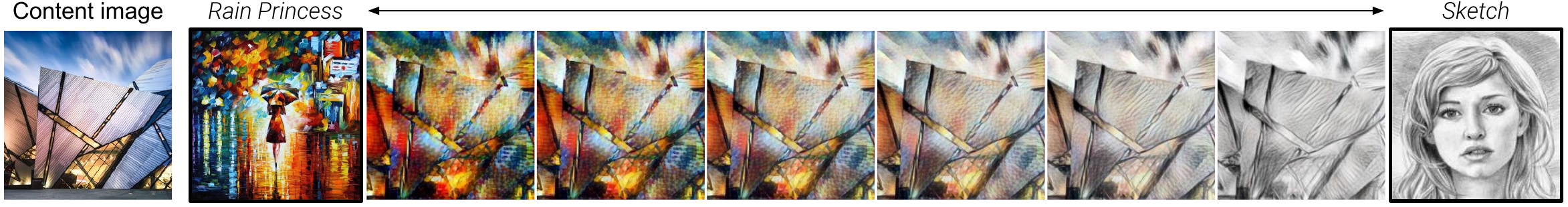}
\caption{Style Interpolation between \emph{Rain Princess} and \emph{Sketch}.}
\label{fig:style_interpolation}
\end{figure*}

\begin{figure*}[t]
\centering
\includegraphics[width=\linewidth]{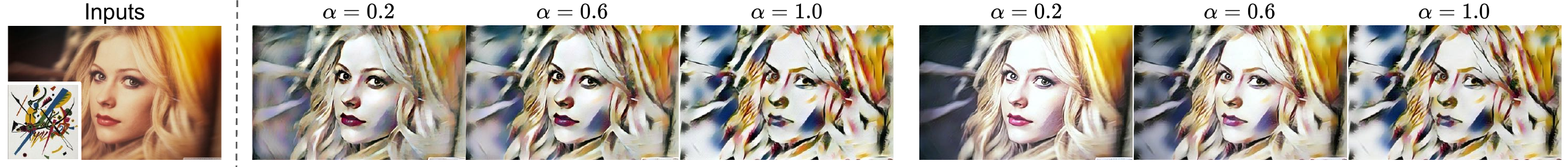}
\caption{Trading off between the content and style images. The left half shows the trading off between the normalized features $\bar{\mathbf{z}}_{cs}$ and $\bar{\mathbf{z}}_{c}$. The right half illustrates the balancing between the content and stylized features $\mathbf{z}_{cs}$ and $\mathbf{z}_c$.}
\label{fig:trade_off_content_and_style}
\end{figure*}

\begin{figure}
\centering
\includegraphics[width=\linewidth]{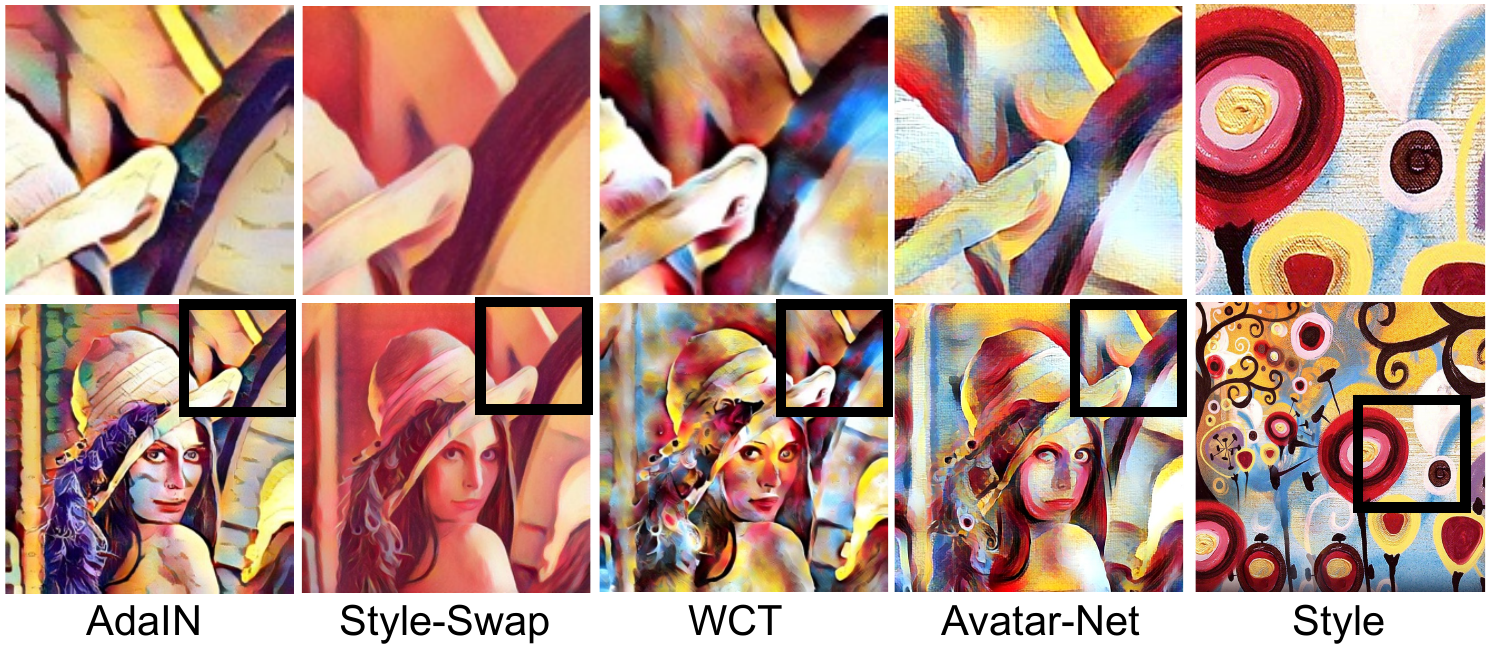}
\caption{Result close-ups. Regions marked by bounding boxes are zoomed in for a better visualization.}
\label{fig:result_close_up}
\end{figure}

\begin{figure}
\centering
\includegraphics[width=\linewidth]{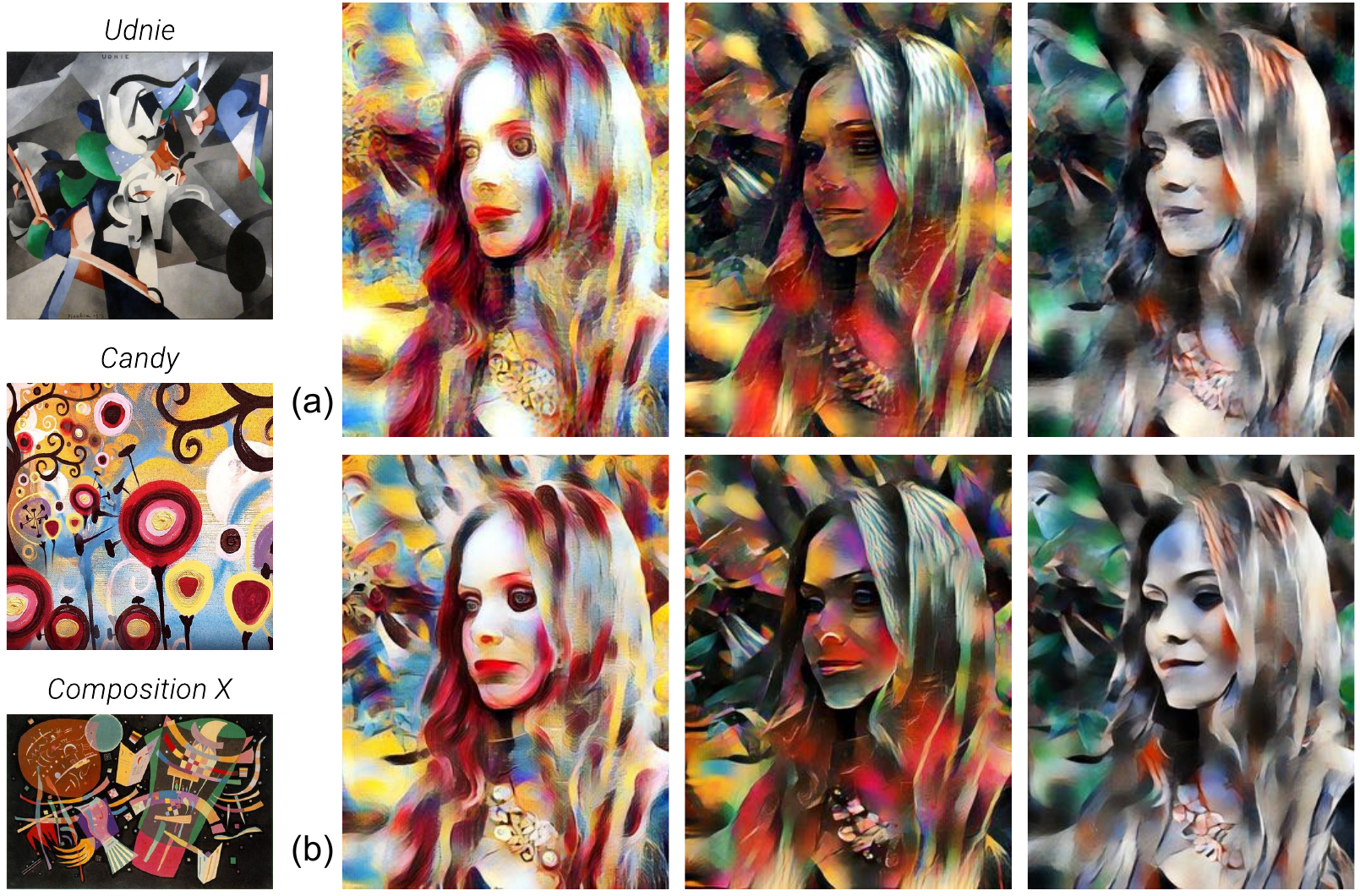}
\caption{Memory reduction. (a) The economical style decorator by sampling the style kernels with a stride of $4$, in which the patch width is $5$. (b) The complete style decorator. The number of style patches in (a) are $\frac{1}{16}$ of that in (b) but the performances are similar except a slight lost of detailed style patterns.}
\label{fig:accelerated_normalized_swap}
\end{figure}

\begin{figure*}[t]
\centering
\includegraphics[width=\linewidth]{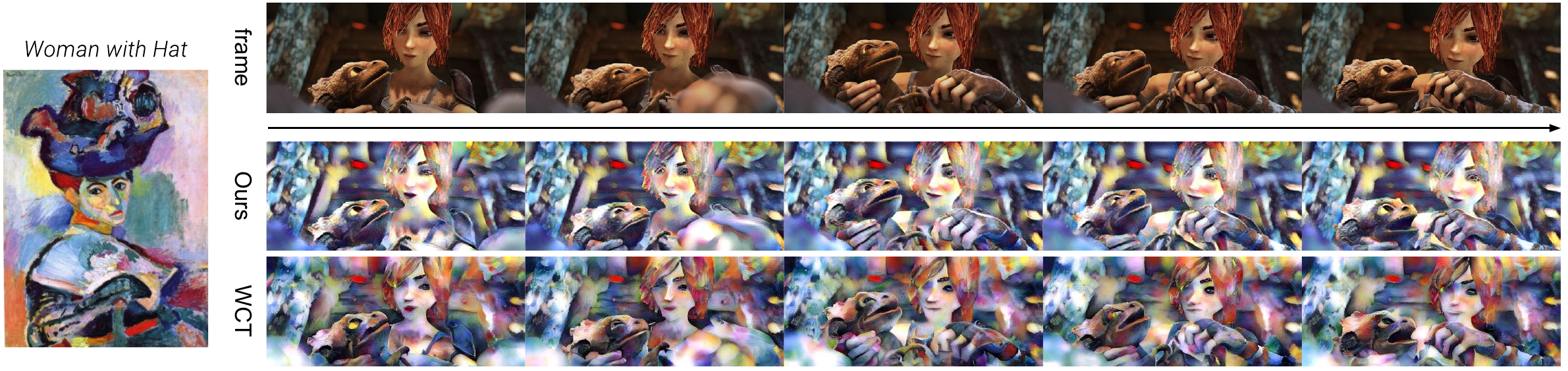}
\caption{Video stylization. A sequence in MPI Sintel dataset~\cite{Butler:ECCV:2012} is stylized by Henri Matisse's \emph{Woman With Hat}. \textbf{Style Decorator}: The stylized patterns in one object keep coherent among adjacent frames. \textbf{WCT}: The results have shape distortions around the face region and contain flickering artifacts among adjacent frames. Please refer to the supplementary materials for a video demonstration.}
\label{fig:video_stylization}
\vspace{-2mm}
\end{figure*}

\subsection{Comparison with Prior Arts}
\label{sub:comparison_with_prior_arts}

To validate the effectiveness of the proposed method, we compare it with two types of zero-shot stylization methods based on (1) iterative optimization~\cite{gatys2016image} and (2) feed-forward network approximation~\cite{huang2017arbitrary,li2017universal,chen2016fast}\footnote{The results by the reference methods are obtained by their public codes with default configurations.}.

\begin{table}[t]
\center
\footnotesize
\begin{tabular}{M{2.4cm}|M{2.3cm}|M{2.3cm}}
\hline
\hline
\multirow{2}{*}{\textbf{Method}} & \multicolumn{2}{c}{\textbf{Execution Time}} \\
\cline{2-3}
& $256\times256$ (sec) & $512\times512$ (sec) \\
\hline
Gatys~\etal~\cite{gatys2016image} & 12.18 & 43.25 \\
AdaIN~\cite{huang2017arbitrary} & 0.053 & 0.11 \\
WCT~\cite{li2017universal} & 0.62 & 0.93 \\
Style-Swap~\cite{chen2016fast} & 0.064 & 0.23 \\
\hline
Ours-ZCA & 0.26 & 0.47 \\
Ours-ZCA-Sampling & 0.24 & 0.32 \\
Ours-AdaIN & 0.071 & 0.28 \\
\hline
\hline
\end{tabular}
\vspace{-3mm}
\caption{Execution time comparison.}
\label{tab:speed_comparison}
\end{table}

\vspace{+1mm}
\noindent\textbf{Qualitative Evaluations.}
The optimization-based approach~\cite{gatys2016image} is able to transfer arbitrary style but the generated results are not stable because the weights trading off the content and style losses are sensitive and the optimization is vulnerable to be stuck onto a bad local minimum, as shown in $4^\text{th}$ and $5^\text{th}$ columns in Fig.~\ref{fig:style_transfer_comparison}.
Both AdaIN~\cite{huang2017arbitrary} and WCT~\cite{li2017universal} holistically adjust the content features to match the second-order statistics of the style features, but AdaIN mimics the channel-wise means and variances and thus provides a suboptimal solution and the learned decoder usually add similar repeated texture patterns for all stylized images.
Although WCT optimally matches the second-order statistics, it cannot always parse the original style patterns and may generate unseen and distorted patterns, for example the background clutters in the $2^\text{nd}$ column and unwanted spiral patterns in the $3^\text{rd}$ column, as well as missed circular patterns in the last column for the style \emph{candy}.
Style-Swap is too rigorous so that the stylized features strictly preserve the content features and only receive low-level style patterns.
This artifact almost occurs in every examples in Fig.~\ref{fig:style_transfer_comparison}.

The proposed method seamlessly reassembles the style patterns according to the semantic spatial distribution of the content image, which works for different kinds of styles, from global hue to local strokes and sketches, as shown in Fig.~\ref{fig:style_transfer_comparison}.
Even though style patterns replace the content information to regenerate the stylized images, necessary content components like human faces, building and skylines are still visually perceived as the combination of the style patterns.

We also provide result close-ups in Fig.~\ref{fig:result_close_up}, the visual result by Avatar-Net receives concrete multi-scale style patterns (e.g., color distribution, brush strokes and circular patterns
in candy image). WCT distorts the brush strokes and circular patterns. AdaIN cannot even keep the color distribution, while style-swap fails in this example.

\vspace{+1mm}
\noindent\textbf{Efficiency.}
In Tab.~\ref{tab:speed_comparison}, we compare the running time with the reference methods.
We implemented these methods on the Tensorflow platform for a fair comparison.
The method by Gatys~\etal~\cite{gatys2016image} is slow since it requires hundreds of forward and backward passes to converge.
WCT~\cite{li2017universal}, AdaIN~\cite{huang2017arbitrary} and Style-Swap~\cite{chen2016fast} are all based on feed-forward networks, in which WCT is relatively slower as it requires several feed-forward passes and the operation of SVD has to be executed in CPU.
Our implemented Style-Swap is efficient in GPU and thus much faster than the reported speed~\cite{chen2016fast}.

The proposed method is more efficient than WCT, because the hourglass architecture enables multi-scale processing in one feed-forward pass.
But the style decorator equipped by ZCA also needs a CPU-based SVD, its speed is thus not comparable to AdaIN and Style-Swap.
We can improve the efficiency by replacing it by AdaIN, in which case the execution time is tremendously reduced to the same level of Style-Swap and AdaIN.
Note that the proposed method can be economical in memory, by randomly (or grid) sampling the style patches in the style decorator procedure.
Due to the local coherence in the style features, we find the resultant stylized images have similar performances as the complete version (in Fig.~\ref{fig:accelerated_normalized_swap}).

\subsection{Applications}
\label{sub:applications}

The flexibility of the Avatar-Net is further manifested by several applications as follows:

\vspace{+1mm}
\noindent\textbf{Trade-off Between Content and Style.}
The degree of stylization can be adjusted by two variants. (1) Adjust the normalized stylized features $\bar{\mathbf{z}}_{cs} \gets \alpha \bar{\mathbf{z}}_{c} + (1-\alpha)\bar{\mathbf{z}}_{cs}, \forall \alpha\in[0,1]$, and then recover the stylized features $\mathbf{z}_{cs}$. (2) Directly adjust the stylized features $\mathbf{z}_{cs} \gets \alpha \mathbf{z}_{c} + (1-\alpha)\mathbf{z}_{cs}$.
In the former variant, $\alpha = 0$ lets the model degrade to the WCT, while $\alpha = 1$ means the image is only reconstructed by the style patterns.
It is harder to control the latter variant due to the magnitude dissimilarities between the content and style features.
We depict these variants in this experiment (in Fig.~\ref{fig:trade_off_content_and_style}), but we do not adjust the low-level style adaptation in the shortcut links, which can be interpolated by~\cite{huang2017arbitrary} to enable multi-scale transfer.

\vspace{+1mm}
\noindent\textbf{Style Interpolation.}
Convexly integrating multiple style images $\{\mathbf{x}_s^k\}_{k=1}^K$ with weights $\{w_k\}_{k=1}^K$ that $\sum_{k=1}^K w_k = 1$, the stylized features are $\mathbf{z}_{cs} = \sum_{k=1}^K w_k \mathbf{z}_{cs}^k$.
The style adaptations in the shortcut links are also extended to a convex combination of the stylized features.
But due to the magnitude dissimilarities among different style features, the interpolated stylization is also affected by their feature magnitudes, as shown in Fig.~\ref{fig:style_interpolation}.

\vspace{+1mm}
\noindent\textbf{Video Stylization.}
In addition to image-based stylization, our model can perform video stylization merely based on per-frame style transfer, as visualized in Fig.~\ref{fig:video_stylization}.
The style decorator module is stable and coherent over adjacent frames since the alignment between feature patches is usually spatially invariant and robust to small content variations.
On the contrary, WCT, as an example, contains severe content distortions and temporal flickering.

\section{Concluding Remarks} 
\label{sec:concluding_remarks}

In this work, we propose a fast and reliable multi-scale zero-shot style transfer method integrating an style decorator for semantic style feature propagation and a hourglass network for multi-scale holistic style adaptation.
Experimental results demonstrate its superiority in generating arbitrary stylized images.
As a future direction, one may replace the projection and reconstruction steps in the style decorator by learnable modules for increased alignment robustness and executing efficiency.

{
\noindent\textbf{Acknowledgments.}
This work is supported in part by SenseTime Group Limited, in part by the General Research Fund through the Research Grants Council of Hong Kong under Grants CUHK14213616, CUHK14206114, CUHK14205615, CUHK419412, CUHK14203015, CUHK14207814, CUHK14202217, and in part by the Hong Kong Innovation and Technology Support Programme Grant ITS/121/15FX.
}

{\small
\bibliographystyle{ieee}
\bibliography{IEEEabrv,egbib}

\begin{thebibliography}{10}\itemsep=-1pt

\bibitem{Butler:ECCV:2012}
D.~J. Butler, J.~Wulff, G.~B. Stanley, and M.~J. Black.
\newblock A naturalistic open source movie for optical flow evaluation.
\newblock In {A. Fitzgibbon et al. (Eds.)}, editor, {\em ECCV}, Part IV, LNCS
  7577, pages 611--625. Springer-Verlag, Oct. 2012.

\bibitem{champandard2016semantic}
A.~J. Champandard.
\newblock Semantic style transfer and turning two-bit doodles into fine
  artworks.
\newblock {\em arXiv preprint arXiv:1603.01768}, 2016.

\bibitem{chen2017coherent}
D.~Chen, J.~Liao, L.~Yuan, N.~Yu, and G.~Hua.
\newblock Coherent online video style transfer.
\newblock In {\em ICCV}, 2017.

\bibitem{chen2017stylebank}
D.~Chen, L.~Yuan, J.~Liao, N.~Yu, and G.~Hua.
\newblock Stylebank: An explicit representation for neural image style
  transfer.
\newblock In {\em CVPR}, 2017.

\bibitem{chen2018stereoscopic}
D.~Chen, L.~Yuan, J.~Liao, N.~Yu, and G.~Hua.
\newblock Stereoscopic neural style transfer.
\newblock {\em CVPR}, 2018.

\bibitem{chen2016fast}
T.~Q. Chen and M.~Schmidt.
\newblock Fast patch-based style transfer of arbitrary style.
\newblock {\em arXiv preprint arXiv:1612.04337}, 2016.

\bibitem{dumoulin2016learned}
V.~Dumoulin, J.~Shlens, and M.~Kudlur.
\newblock A learned representation for artistic style.
\newblock In {\em ICLR}, 2017.

\bibitem{efros2001image}
A.~A. Efros and W.~T. Freeman.
\newblock Image quilting for texture synthesis and transfer.
\newblock In {\em Proceedings of the 28th annual conference on Computer
  graphics and interactive techniques}, pages 341--346. ACM, 2001.

\bibitem{elad2017style}
M.~Elad and P.~Milanfar.
\newblock Style transfer via texture synthesis.
\newblock {\em {IEEE} Trans. Image Process.}, 26(5):2338--2351, 2017.

\bibitem{gatys2015texture}
L.~A. Gatys, A.~S. Ecker, and M.~Bethge.
\newblock Texture synthesis using convolutional neural networks.
\newblock In {\em NIPS}, pages 262--270, 2015.

\bibitem{gatys2016image}
L.~A. Gatys, A.~S. Ecker, and M.~Bethge.
\newblock Image style transfer using convolutional neural networks.
\newblock In {\em CVPR}, pages 2414--2423, 2016.

\bibitem{gupta2017characterizing}
A.~Gupta, J.~Johnson, A.~Alahi, and L.~Fei-Fei.
\newblock Characterizing and improving stability in neural style transfer.
\newblock In {\em ICCV}, 2017.

\bibitem{hertzmann2001image}
A.~Hertzmann, C.~E. Jacobs, N.~Oliver, B.~Curless, and D.~H. Salesin.
\newblock Image analogies.
\newblock In {\em Proceedings of the 28th annual conference on Computer
  graphics and interactive techniques}, pages 327--340. ACM, 2001.

\bibitem{huang2017arbitrary}
X.~Huang and S.~Belongie.
\newblock Arbitrary style transfer in real-time with adaptive instance
  normalization.
\newblock In {\em ICCV}, 2017.

\bibitem{johnson2016perceptual}
J.~Johnson, A.~Alahi, and L.~Fei-Fei.
\newblock Perceptual losses for real-time style transfer and super-resolution.
\newblock In {\em ECCV}, pages 694--711. Springer, 2016.

\bibitem{kessy2017optimal}
A.~Kessy, A.~Lewin, and K.~Strimmer.
\newblock Optimal whitening and decorrelation.
\newblock {\em The American Statistician}, (just-accepted), 2017.

\bibitem{kyprianidis2013state}
J.~E. Kyprianidis, J.~Collomosse, T.~Wang, and T.~Isenberg.
\newblock State of the ``art'': A taxonomy of artistic stylization techniques
  for images and video.
\newblock {\em {IEEE} Trans. Vis. Comput. Graphics}, 19(5):866--885, 2013.

\bibitem{li2016combining}
C.~Li and M.~Wand.
\newblock Combining markov random fields and convolutional neural networks for
  image synthesis.
\newblock In {\em CVPR}, pages 2479--2486, 2016.

\bibitem{li2016precomputed}
C.~Li and M.~Wand.
\newblock Precomputed real-time texture synthesis with markovian generative
  adversarial networks.
\newblock In {\em ECCV}, pages 702--716. Springer, 2016.

\bibitem{li2017diversified}
Y.~Li, C.~Fang, J.~Yang, Z.~Wang, X.~Lu, and M.-H. Yang.
\newblock Diversified texture synthesis with feed-forward networks.
\newblock In {\em CVPR}, 2017.

\bibitem{li2017universal}
Y.~Li, C.~Fang, J.~Yang, Z.~Wang, X.~Lu, and M.-H. Yang.
\newblock Universal style transfer via feature transforms.
\newblock In {\em NIPS}, 2017.

\bibitem{li2017Demystifying}
Y.~Li, N.~Wang, J.~Liu, and X.~Hou.
\newblock Demystifying neural style transfer.
\newblock In {\em IJCAI}, pages 2230--2236, 2017.

\bibitem{liao2017visual}
J.~Liao, Y.~Yao, L.~Yuan, G.~Hua, and S.~B. Kang.
\newblock Visual attribute transfer through deep image analogy.
\newblock {\em arXiv preprint arXiv:1705.01088}, 2017.

\bibitem{lin2014microsoft}
T.-Y. Lin, M.~Maire, S.~Belongie, J.~Hays, P.~Perona, D.~Ramanan,
  P.~Doll{\'a}r, and C.~L. Zitnick.
\newblock Microsoft coco: Common objects in context.
\newblock In {\em ECCV}, pages 740--755. Springer, 2014.

\bibitem{luan2017deep}
F.~Luan, S.~Paris, E.~Shechtman, and K.~Bala.
\newblock Deep photo style transfer.
\newblock {\em CVPR}, 2017.

\bibitem{ruder2016artistic}
M.~Ruder, A.~Dosovitskiy, and T.~Brox.
\newblock Artistic style transfer for videos.
\newblock In {\em German Conference on Pattern Recognition}, pages 26--36.
  Springer, 2016.

\bibitem{shen2017meta}
F.~Shen, S.~Yan, and G.~Zeng.
\newblock Meta networks for neural style transfer.
\newblock {\em arXiv preprint arXiv:1709.04111}, 2017.

\bibitem{shih2014style}
Y.~Shih, S.~Paris, C.~Barnes, W.~T. Freeman, and F.~Durand.
\newblock Style transfer for headshot portraits.
\newblock {\em TOG}, 2014.

\bibitem{simonyan2014very}
K.~Simonyan and A.~Zisserman.
\newblock Very deep convolutional networks for large-scale image recognition.
\newblock In {\em ICLR}, 2015.

\bibitem{ulyanov2016texture}
D.~Ulyanov, V.~Lebedev, A.~Vedaldi, and V.~S. Lempitsky.
\newblock Texture networks: Feed-forward synthesis of textures and stylized
  images.
\newblock In {\em ICML}, pages 1349--1357, 2016.

\bibitem{wang2017zm}
H.~Wang, X.~Liang, H.~Zhang, D.-Y. Yeung, and E.~P. Xing.
\newblock Zm-net: Real-time zero-shot image manipulation network.
\newblock {\em arXiv preprint arXiv:1703.07255}, 2017.

\bibitem{wilmot2017stable}
P.~Wilmot, E.~Risser, and C.~Barnes.
\newblock Stable and controllable neural texture synthesis and style transfer
  using histogram losses.
\newblock {\em arXiv preprint arXiv:1701.08893}, 2017.

\bibitem{zhang2017multistyle}
H.~Zhang and K.~Dana.
\newblock Multi-style generative network for real-time transfer.
\newblock {\em arXiv preprint arXiv:1703.06953}, 2017.

\end{thebibliography}
}

\begin{figure*}
\includegraphics[width=\linewidth]{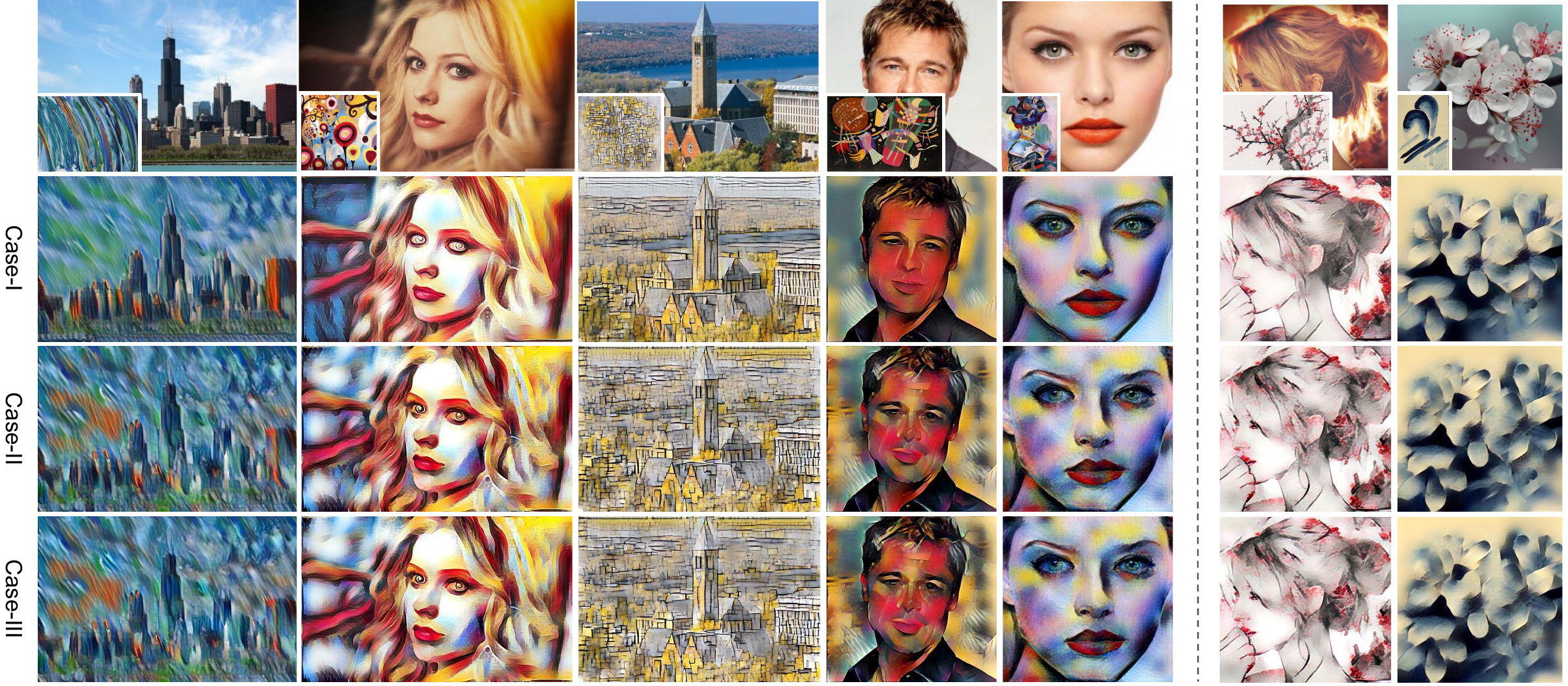}
\captionof{figure}{Comparison among different projection and reconstruction methods. Case-I by AdaIN usually preserves more content patterns and thus receives fewer style patterns. Case-II and Case-III always have very similar stylization results. But in some cases these methods have similar performances, such as the right two columns that these methods have comparable stylization results.}
\label{fig:comparison_projection_and_reconstruction}
\end{figure*}

\newpage

\section{Appendix}

\subsection{Ablation Study of Style Decorator}
\label{sec:ablation_study_of_style_decorator}

\subsubsection{Projection and Reconstruction}
\label{sub:projection_and_reconstruction}

We examine three types of projection and reconstruction transforms in the proposed style decorator.

\vspace{+1mm}
\noindent\textbf{Case-I: AdaIN~\cite{huang2017arbitrary}.~}
It transfers the channel-wise statistics ($\boldsymbol\mu(\mathbf{z}_s)$ and $\boldsymbol\sigma(\mathbf{z}_s)$) to the instance normalized features $\bar{\mathbf{z}}_c$.

\noindent\textbf{Case-II: ZCA~\cite{kessy2017optimal} for $\mathcal{C}(\mathbf{z})$.~}
It is the method used in this paper.
By singular value decomposing the covariance matrix $\mathcal{C}(\mathbf{z}) = \frac{1}{N}\sum_{n=1}^N (\mathbf{z}_n - \boldsymbol\mu(\mathbf{z}))(\mathbf{z}_n - \boldsymbol\mu(\mathbf{z}))^\top$ as $\mathbf{U}\boldsymbol\Sigma\mathbf{V}^\top$, the projection and reconstruction kernels are
\begin{equation}
\mathbf{W} = \mathbf{U}(\boldsymbol\Sigma + \varepsilon\mathbf{I})^{-1}\mathbf{U}^\top~\text{and}~\mathbf{C} = \mathbf{U}(\boldsymbol\Sigma+\varepsilon\mathbf{I})\mathbf{U}^\top,
\end{equation}
where $\varepsilon$ is a small value avoiding bad matrix inversion.
$N$ is the number of elements and $\mathbf{z}_n \in \mathbb{R}^{C}$.
The normalized features $\bar{\mathbf{z}}$ are uniformly distributed in the standard Gaussian space, and each element $\bar{\mathbf{z}}_n$ is oriented with the same direction as the original feature $\mathbf{z}_n - \boldsymbol\mu(\mathbf{z})$ thus inherently encodes the feature variations of $\mathbf{z}$ in the unit sphere.

\begin{figure*}
\centering
\includegraphics[width=\linewidth]{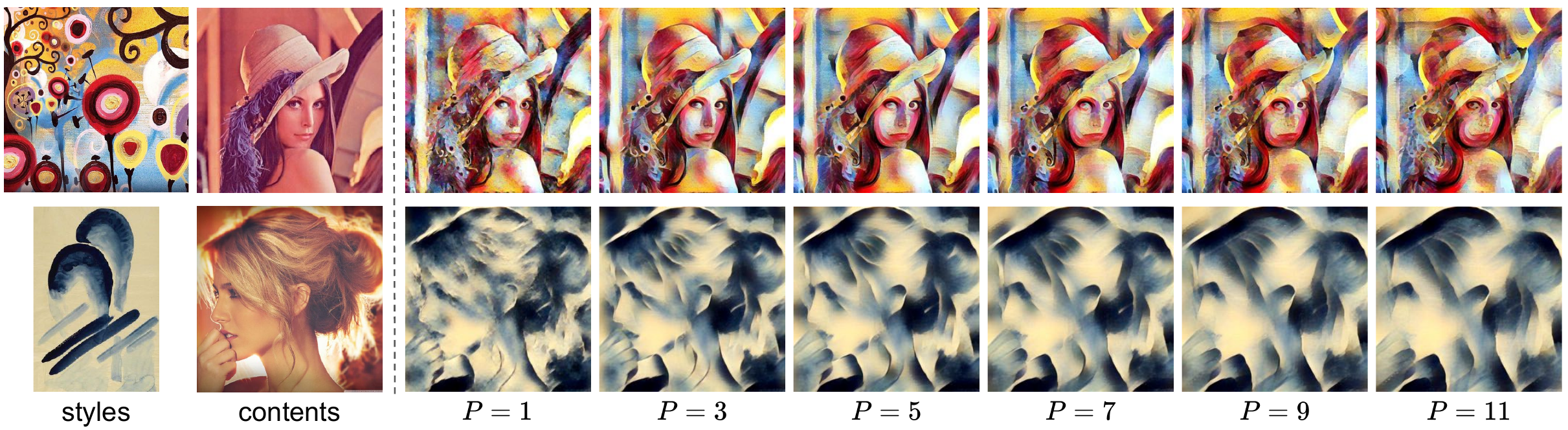}
\vspace{-5mm}
\caption{Varying patch sizes affecting the scales of the transferred style patterns. A larger patch size leads to more global style patterns. When $P=11$, its performance is still better than WCT~\cite{li2017universal}, as the results of \emph{Lena} shown in Fig.~6 in the main article.}
\label{fig:comparison_patch_size}
\end{figure*}

\begin{figure*}
\centering
\includegraphics[width=\linewidth]{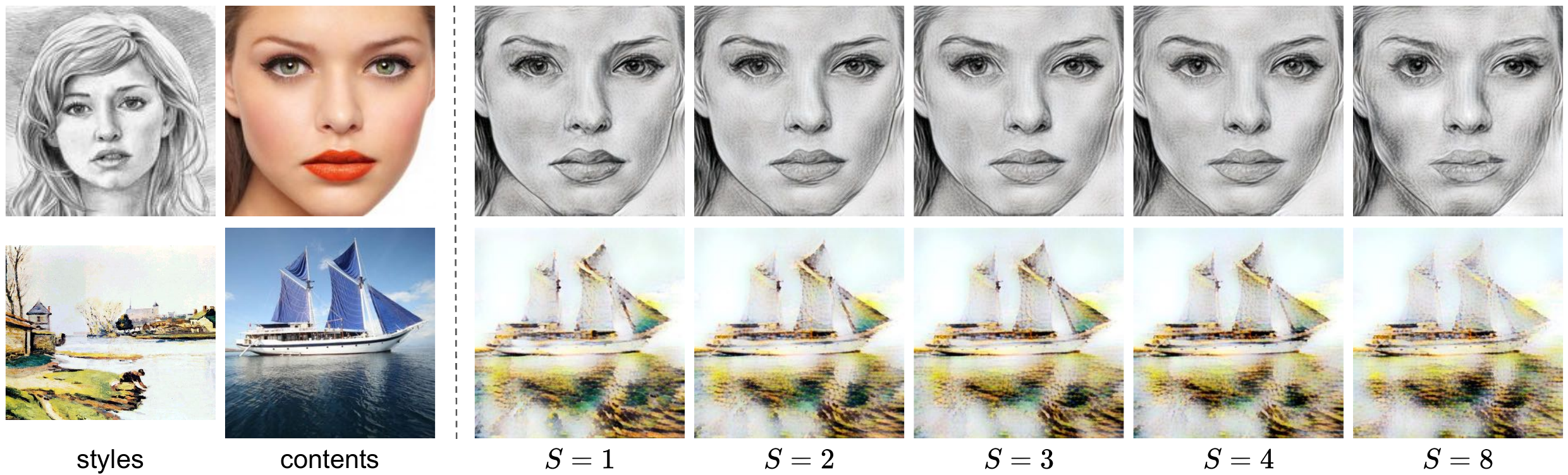}
\vspace{-5mm}
\caption{The comparison among different sampling densities in the style decorator. The density is controlled by the stride in extracting the style patches. With the increasing of the stride, fewer style patterns are kept in the stylized images. Note that the patch size applied is $P=5$. We also find that the small stride may not degrade the results with apparent visual artifacts. But larger strides (such as $S=8$) will reduce quite a lot of style patterns since this stride skips a lot of style patterns.}
\label{fig:comparison_patch_sampling}
\end{figure*}

\noindent\textbf{Case-III: ZCA for $\mathcal{G}(\mathbf{z})$.~}
Alternately, ZCA can be performed on the Gram matrix $\mathcal{G}(\mathbf{z}) = \frac{1}{N}\sum_{n=1}^N \mathbf{z}_n\mathbf{z}_n^\top$ instead of the covariance matrix $\mathcal{C}(\mathbf{z})$.
The normalized features $\bar{\mathbf{z}}$ also follow a standard Gaussian distribution, but they are oriented as the original features $\mathbf{z}_n$.

\vspace{+1mm}
By applying these methods into our style decorator modules, as shown in Fig.~\ref{fig:comparison_projection_and_reconstruction}, we find that AdaIN cannot dispel all the texture characteristics of the content features, and thus the matched style patches have a limited diversity.
Generally, Case-I still sometimes performs similarly as Case-II and Case-III, we may apply Case-I instead of Case-II for the sake of efficiency in real-time applications.

Case-II and Case-III have very similar performance with each other, which means the patch matching by cosine distance is somewhat robust to these normalization strategies.
However, we need to mention that WCT~\cite{li2017universal} requires Case-II for a satisfactory result since Case-III will bias the normalized features to the direction of the mean features, thus there are larger orientation discrepancy between the normalized content features and the normalized style features.

\subsubsection{Matching and Reassembling}
\label{sub:matching_and_reassembling}

\noindent\textbf{Patch Size.~}
The increasing of the patch size in the matching and reassembling step tends to increase the scale of the style patterns in the stylized images.
As visualized in Fig.~\ref{fig:comparison_patch_size}, more global style patterns are presented in the stylized images with larger distortions towards the content information, when larger patch size is applied.
Specially, when the patch size $P=11$, the stylized images have a similar scale of style patterns as WCT, but the results are much more vivid.

\vspace{+1mm}
\noindent\textbf{Patch Sampling.~}
The sampling density of the style patches controls the diversity and completeness of the style patterns in the stylized images.
We control the density by extracting the style patches with a controllable stride. 
As visualized in Fig.~\ref{fig:comparison_patch_sampling}, the fewer samples are stored in the style kernel, the fewer style patterns are recovered in the stylized images.
But we find that smaller strides may just have a slight performance degradation (\ie, the stride $S$ is not larger than $P-1$), which means that we may reduce the memory cost and computational complexity in this step by a stride $1 < S < P$.

\subsection{Exemplar Stylization Results}
\label{sec:exemplar_stylization_results}

\begin{figure*}
\centering
\subfigure[Exemplar Set $1$]{\includegraphics[width=\linewidth]{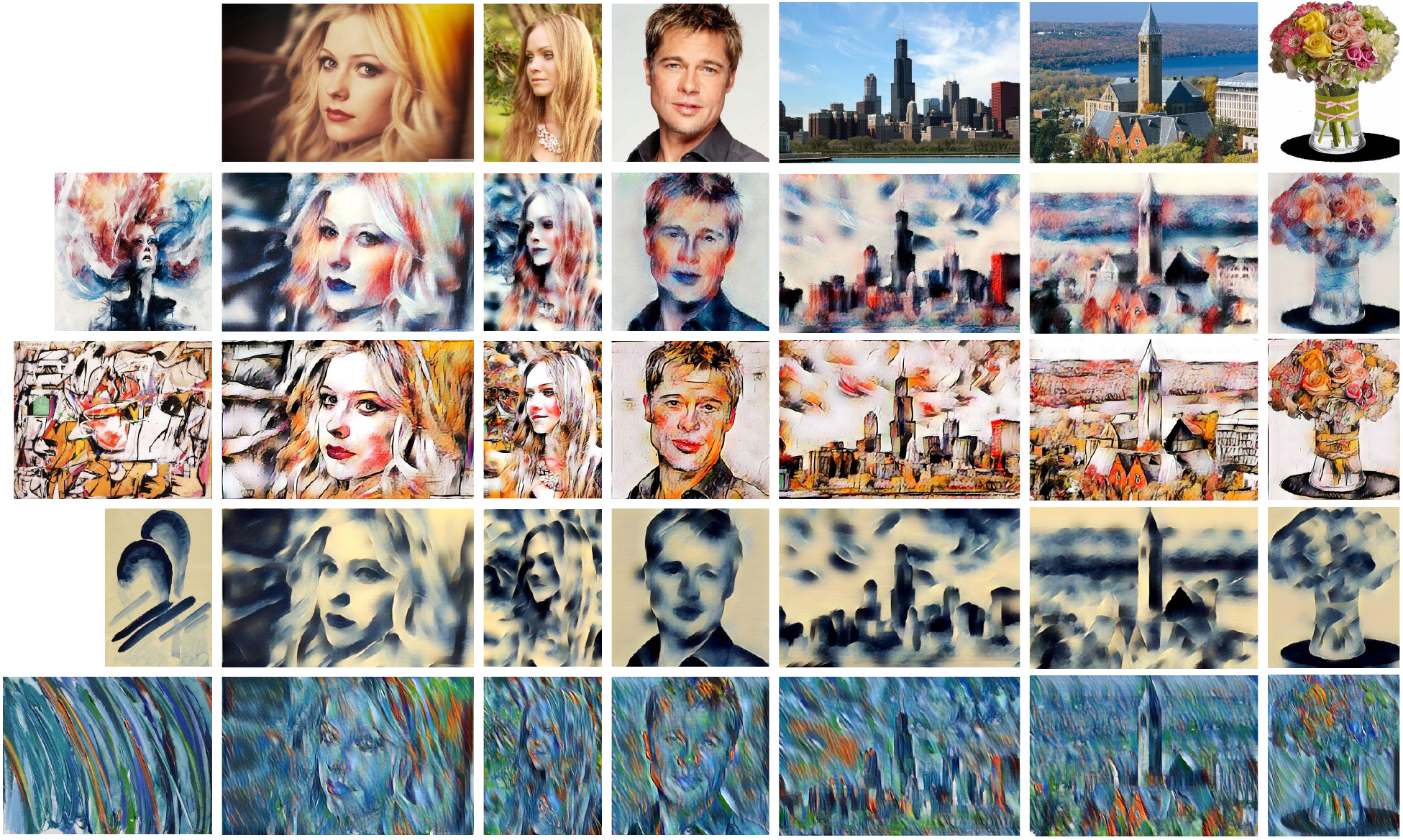}}
\subfigure[Exemplar Set $2$]{\includegraphics[width=\linewidth]{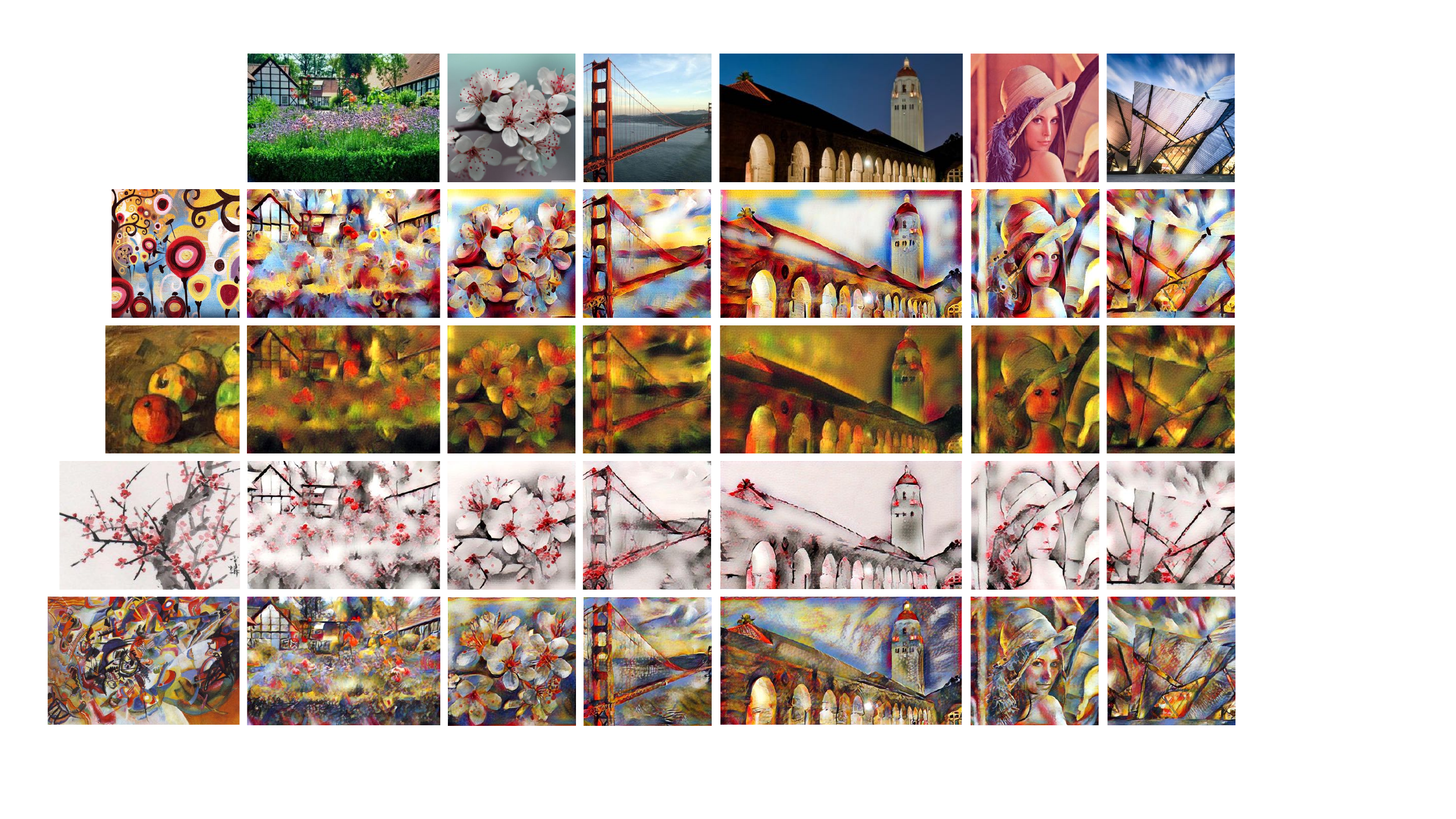}}
\vspace{-3mm}
\caption{Exemplar stylization results.}
\label{fig:examples_1}
\end{figure*}

\begin{figure*}
\centering
\includegraphics[width=\linewidth]{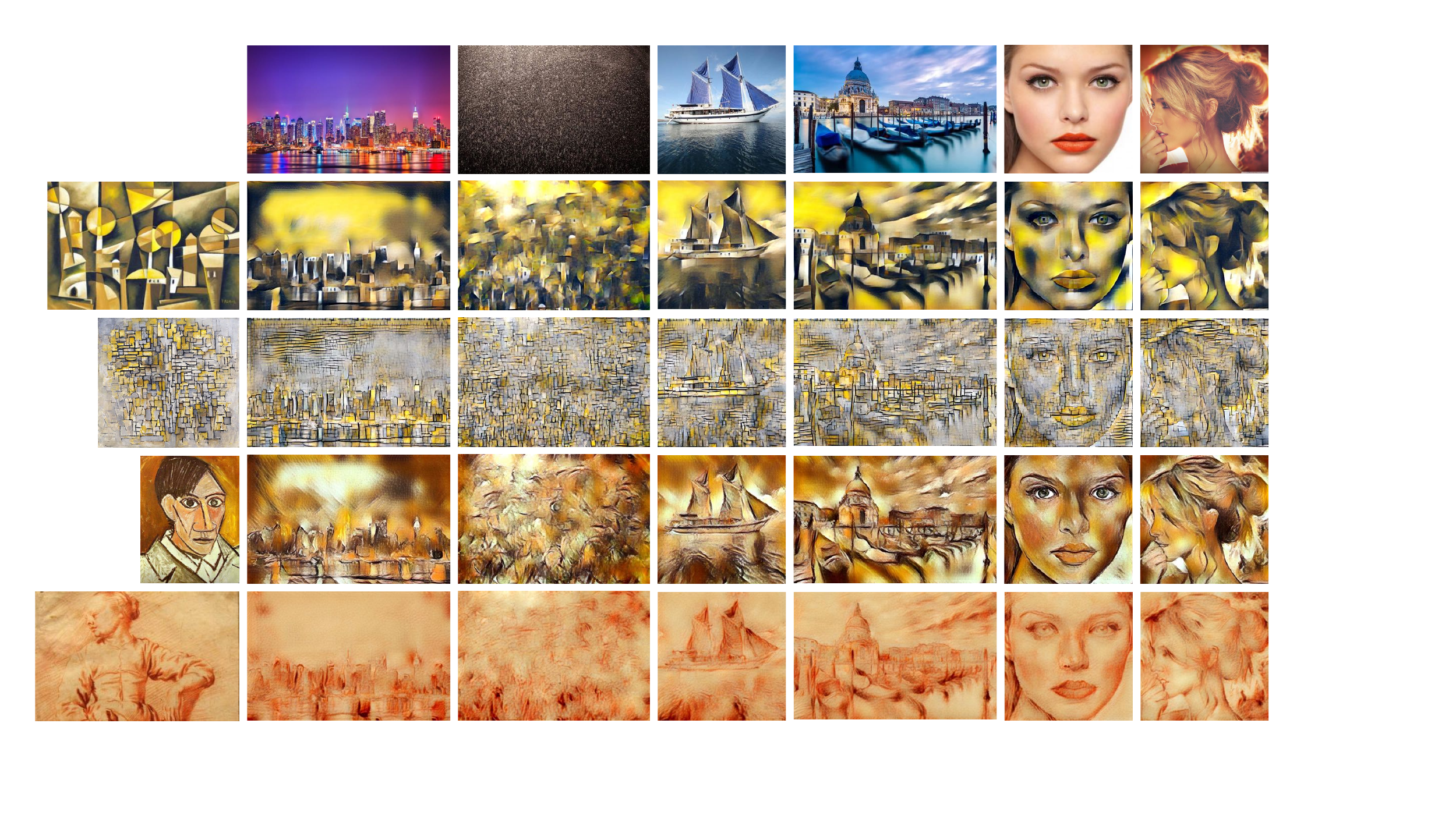}
\caption{Exemplar stylization results for exemplar set $3$.}
\label{fig:examples_2}
\end{figure*}

In this part, we show some additional stylization results by the proposed methods, as visualized in Fig.~\ref{fig:examples_1} and \ref{fig:examples_2}.
If not specifically stated, the patch size $P=5$ and the blending weight is $\alpha = 0.8$ for the proposed style transfer both in the main article and this supplementary material.

\subsection{Video Stylization}
\label{sec:video_stylization}

We also present several demos comparing the stylized videos generated by the proposed method.
Please refer to the YouTube link: \url{https://youtu.be/amaeqbw6TeA}

\end{document}